\documentclass[letterpaper]{article} % DO NOT CHANGE THIS
\usepackage{aaai2027}
\usepackage[table]{xcolor}
\usepackage{subcaption}
\usepackage[hyphens]{url}  % DO NOT CHANGE THIS
\usepackage{graphicx} % DO NOT CHANGE THIS
\usepackage{natbib}  % DO NOT CHANGE THIS AND DO NOT ADD ANY OPTIONS TO IT
\usepackage{caption} % DO NOT CHANGE THIS AND DO NOT ADD ANY OPTIONS TO IT
\usepackage{algorithm}
\usepackage{algorithmic}
\usepackage{pifont}
\usepackage{algorithm}
\usepackage{algorithmic}
\usepackage{multirow}
\usepackage{graphicx}
\usepackage{amsmath}
\usepackage{booktabs}
\usepackage{pifont}
\usepackage{array}
\usepackage{bbding}
\usepackage{amssymb}
\usepackage{newfloat}
\usepackage{listings}
\DeclareCaptionStyle{ruled}{labelfont=normalfont,labelsep=colon,strut=off} % DO NOT CHANGE THIS
\floatstyle{ruled}
\newfloat{listing}{tb}{lst}{}
\floatname{listing}{Listing}

\usepackage{booktabs}

\nocopyright 
\title{SpikingMOT: A Spike-Driven Multi-Object Tracker}
\author{
    Yiding Sun\textsuperscript{\rm 1,2},
    Xiangyang Yang\textsuperscript{\rm 1},
    Dongxu Zhang\textsuperscript{\rm 1},
    Qirui Wang\textsuperscript{\rm 1},
    Zijie Xu\textsuperscript{\rm 2},
    Wenxuan Liu\textsuperscript{\rm 2},
    Shuiwang Li\textsuperscript{\rm 3},
    Jihua Zhu\textsuperscript{\rm 1},
    Zhaofei Yu\textsuperscript{\rm 2},
    Tiejun Huang\textsuperscript{\rm 2}
}
\affiliations{
    \textsuperscript{\rm 1}Xi'an Jiaotong University\\
    \textsuperscript{\rm 2}Peking University\\
    \textsuperscript{\rm 3}Guilin University of Technology\\
}

\begin{document}

\maketitle

\begin{abstract}
Multi-object tracking (MOT) plays a fundamental role in visual perception, where accurate trajectory prediction is essential for reliable target association under complex motion patterns. Recent trackers have improved motion modeling with densely activated artificial neural networks, yet they largely overlook whether such dense responses are necessary for trajectory prediction. In this paper, we formulate activation sparsity preference (ASP) by tackling two key questions: \ding{172} How can we identify a model architecture that appropriately and formally explains ASP, and \ding{173} How can we translate this explanation into competitive tracking performance. Theoretical analysis shows that sparse gating is no worse than state-independent dropout under the same activation rate. Based on this insight, SpikingMOT is proposed as a spike-driven tracker that adaptively models sparse trajectory dynamics with spiking neural networks (SNNs). Specifically, SpikingMOT decomposes each trajectory state into pseudo-trajectory bases and uses the current prediction error to calibrate the posterior for next-frame prediction. With this brain-inspired loop, SpikingMOT achieves state-of-the-art performance in extensive experiments, \textit{e.g.} 74.9 HOTA on SportsMOT and 56.5 HOTA on DanceTrack, while reducing the parameters and energy by 72\% and 86.7\%, respectively. These results bring SNNs into MOT, opening a promising direction for efficient tracking.
\end{abstract}


\section{Introduction}
Multi-object tracking (MOT) is a fundamental task in computer vision, with broad applications in surveillance, autonomous driving~\cite{sun2026tri,sun2026align}, and robotics~\cite{guo2026mantis,wang2026pointrft}. Under this topic, Tracking-by-detection (TBD) has become the dominant paradigm due to its concise design, which typically consists of two steps: 1) using an off-the-shelf detector to obtain object bounding boxes, and 2) tracking targets across frames with motion or appearance cues. Over the past decade, these two components have often been optimized separately while still complementing each other within a complete tracking system.

Although classical TBD trackers instantiate this prior using filter-based methods~\cite{zhang2022bytetrack,bewley2016sort}, such approaches often struggle to maintain reliable associations in complex scenarios~\cite{sun2022dancetrack}.
To overcome these challenges, learning-based methods, such as MambaTrack~\cite{xiao2024mambatrack}, have emerged as a promising alternative and attracted considerable attention. 

\begin{figure}[t]
\centering
\includegraphics[width=\linewidth]{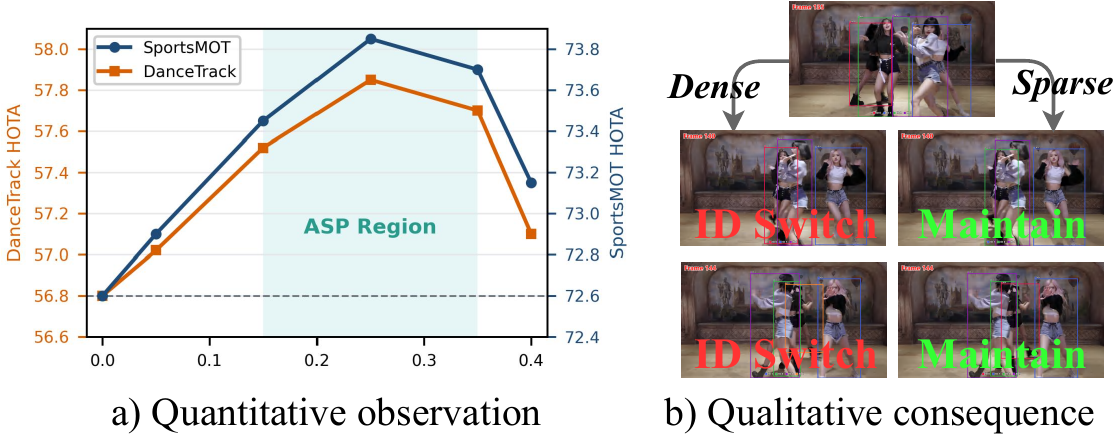} % Reduce the figure size so that it is slightly narrower than the column.
\caption{\textbf{Empirical evidence for ASP.}
(a) Tracking performance under progressively reduced activation rates in MambaTrack with all protocols fixed. Moderate suppression consistently improves HOTA within a nontrivial range.
(b) Tracking cases showing that uniformly dense responses can amplify redundant motion cues and cause inaccurate prediction or identity switches, whereas moderate suppression preserves informative motion responses.}
\label{fig1}
\end{figure}

% Instead of imposing hand-crafted assumptions about regular motion patterns, MambaTrack employs an autoregressive predictor~\cite{Gu_2024_COLM} to learn diverse motion dynamics from historical trajectories. 
% This formulation provides a flexible mechanism for modeling temporal dependencies, thereby maintaining strong performance. Follow-up studies further strengthen this line through trajectory representation~\cite{huang2025mambamot}, adaptive association~\cite{khanna2025sportmamba}, and flow decoding~\cite{hu2026trackssm}.

Instead of relying on hand-crafted assumptions about regular motion, MambaTrack employs an autoregressive predictor~\cite{Gu_2024_COLM} to learn motion dynamics from historical trajectories. Subsequent studies extend this paradigm through improved trajectory representations~\cite{huang2025mambamot}, adaptive association~\cite{khanna2025sportmamba}, and flow-based decoding~\cite{hu2026trackssm}. 

\begin{figure}[t]
\centering
\includegraphics[width=.9\linewidth]{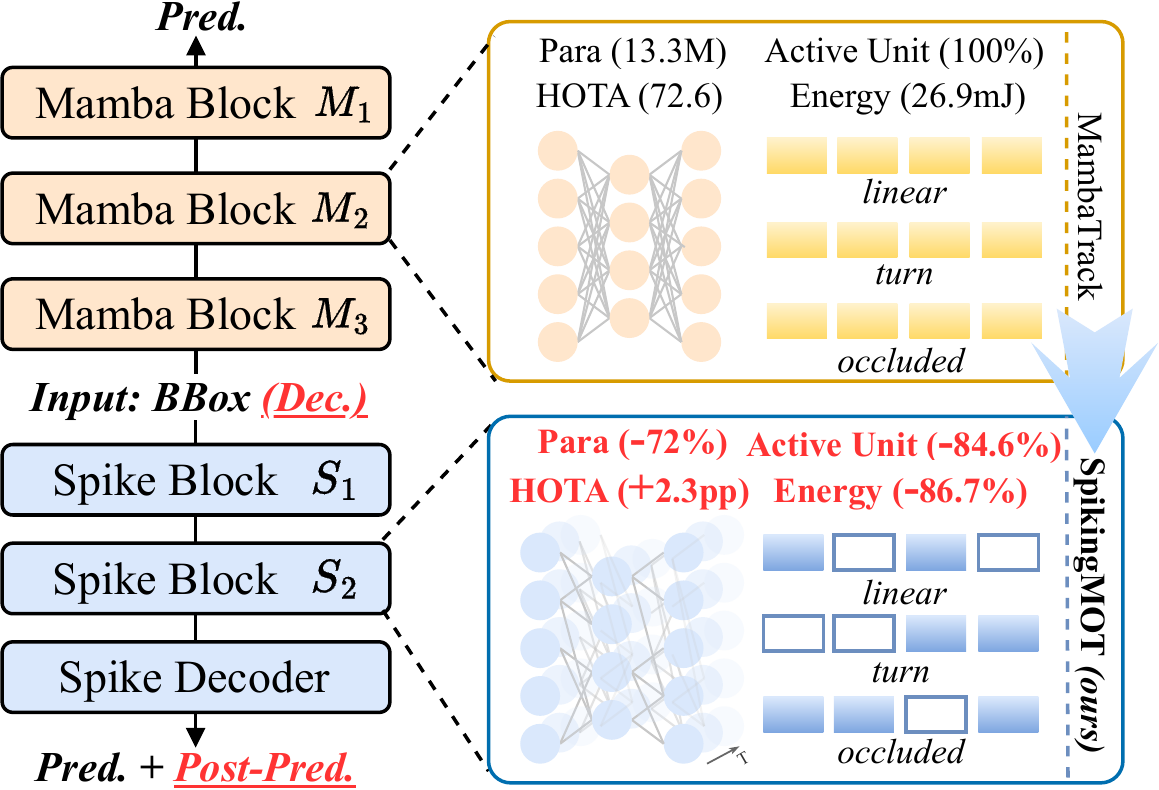} % Reduce the figure size so that it is slightly narrower than the column.
\caption{Inspired by ASP, SpikingMOT replaces dense ANN-based trajectory modeling with sparse spiking dynamics, enabling efficient motion prediction. \textcolor[HTML]{FF3333}{Red} text highlight the characteristics and advantages of SpikingMOT.}
\label{fig2}
\end{figure}

Nevertheless, these methods still rely on densely activated artificial neural networks (ANNs), potentially introducing redundant motion responses and unnecessary computation. To examine whether dense activation is essential for trajectory prediction, we conduct a controlled dropout study on MambaTrack while keeping the remaining tracking pipeline unchanged. Fig.~\ref{fig1} indicates moderate suppression not only preserves performance but improves it within a nontrivial range. We term this tendency the \textbf{Activation Sparsity Preference (ASP)}, suggesting that sparse activations are not merely tolerated by trajectory prediction, but can actively benefit it.

% Despite these advances, most prior arts still rely on densely activated artificial neural networks (ANN),  which introduce redundant motion responses and extra computational overhead. This raises a largely overlooked question: \textbf{Are dense ANN activations truly necessary for trajectory prediction?} To investigate this issue, we conduct a controlled dropout study on MambaTrack, progressively suppressing its activations while keeping the remaining tracking pipeline unchanged. As illustrated in Fig.~\ref{fig1}, moderate dropout does not merely preserve tracking performance, but can further improve it within a certain range. This phenomenon suggests that sparse activations are not merely tolerated by trajectory prediction, but can actively benefit it.

Spiking neural networks (SNNs) perform event-driven computation through sparse binary spikes and stateful membrane dynamics, making them a natural architectural candidate for realizing ASP. However, directly replacing an ANN motion predictor with a generic SNN backbone remains insufficient. Sparse firing determines whether a response is activated, but does not specify how heterogeneous motion patterns should be organized into selectable hypotheses or how these hypotheses should be updated after new observations. Effective tracking therefore requires two task-specific capabilities. One provides a structured motion representation for sparse selection, while the other adapts the selected responses using online observations.

Accordingly, we first show that state-conditioned sparse gating is no worse than state-independent dropout under the same activation rate \underline{\textit{(cf. Supp A)}}. Based on this insight, we propose SpikingMOT, the first spike-driven tracker for MOT. Specifically, building upon SNNs, we develop a brain-inspired strategy tailored to ASP, which consists of two alternating optimization steps: decomposition and calibration. Unlike a generic SNN, the decomposition step parses each trajectory box into multiple pseudo-trajectories, mimicking population coding in the brain~\cite{pouget2000information}, where a continuous motion state is represented by a distributed set of selectively activated neural populations. Prediction-error calibration then connects two consecutive prediction steps. After association at frame $f$, it measures the residual between the predicted box and its matched detection. This residual adjusts the contribution of each motion basis, and the calibrated weights are carried forward to predict frame $f+1$. In this way, the first component provides explicit motion candidates for sparse selection, while the second corrects that selection using newly observed motion. By iteratively performing these two steps, SpikingMOT adaptively learns the motion distribution of each scene, thereby releasing the full potential of SNNs for trajectory prediction in Fig.~\ref{fig2}. Our main contributions are fourfold.

\begin{itemize}
    \item We explore the ASP phenomenon within MOT trajectory prediction and provide a biological plausibility from population coding and predictive coding.
    \item Building upon this insight, we propose SpikingMOT, the first spike-driven tracker for MOT, whose spatial-temporal dynamics provide adaptive sparse activations.
    \item Derived from the framework, we present a brain-inspired strategy for ASP, which introduces a pseudo-trajectory decomposition and a prediction-error calibration.
    \item Extensive experiments show that SpikingMOT exhibits superior performance, \textit{e.g.}, it attains state-of-the-art HOTA of 74.9 on SportsMOT and 56.5 on DanceTrack, while reducing 72\% parameters and 86.7\% energy.
\end{itemize}

\section{Related Works}
\subsection{Multi-Object Tracking}
MOT methods can be broadly grouped into two paradigms: TBD and joint detection and embedding (JDE). JDE methods seek to fuse detection and tracking into a single network, whereas TBD methods typically decouple the pipeline into multiple modules. As an early attempt, Tracktor~\cite{bergmann2019tracktor} uses the regression head of Faster R-CNN~\cite{ren2015fasterrcnn} to predict object locations in the next frame, but it struggles under low-frame-rate scenarios. JDE~\cite{wang2020jde} extends a YOLO-based~\cite{redmon2016yolo} detector with an additional ReID branch to extract identity embeddings, yet the detection and embedding objectives are considered to compete with each other during optimization. To alleviate this issue, FairMOT~\cite{zhang2021fairmot} builds upon the anchor-free detector CenterNet~\cite{zhou2019objects} and introduces a more balanced design, leading to improved tracking performance. With the advent of DETR~\cite{carion2020detr}, several methods have explored its cross-frame query mechanism for implicit association. TransTrack~\cite{sun2020transtrack}, TrackFormer~\cite{meinhardt2022trackformer} and MOTR~\cite{zeng2022motr} are among the first to bridge detection and tracking within this framework. Subsequent approaches further improve the paradigm from the perspectives of memory encoding~\cite{cai2022memot} and background awareness. While the community has continuously advanced detector architectures, limited attention has been paid to the design of the tracker itself. In this paper, SpikingMOT shifts the focus to the tracking stage and pioneers the first spike-driven tracker for MOT.

\subsection{Motion Models in MOT}
Motion is an inherent property of objects. Early classical MOT trackers, such as SORT~\cite{bewley2016sort} and DeepSORT~\cite{wojke2017deepsort}, rely on the Kalman filter (KF)~\cite{kalman1960filter}  to predict each target's inter-frame displacement. Later methods along this line refine the KF from different perspectives. For instance, OC-SORT~\cite{cao2023ocsort} incorporates nonlinear motion cues, while Deep OC-SORT~\cite{maggiolino2023deepocsort} adaptively incorporates appearance cues into the observation-centric association process. Despite these advances, filter-based methods still depend heavily on predefined hyperparameters and often fail to generalize to real-world scenarios. As a successor, learning-based methods use data-driven models to mitigate the limitations. These approaches learn from an object's historical trajectory and predict its future state with neural networks. MotionTrack~\cite{xiao2024motiontrack} demonstrates the feasibility of this direction with a Transformer architecture. DiffMOT~\cite{lv2024diffmot} and DiffusionTrack~\cite{luo2024diffusiontrack} reformulate motion prediction as a generative task and achieve strong performance. MambaTrack~\cite{xiao2024mambatrack} and its variants further exploit hidden states for data diversity. Nevertheless, existing models rarely question whether dense motion modeling is necessary for trajectory prediction. To investigate this question, SpikingMOT explores the ASP phenomenon and introduces sparse spikes as an effective solution.

\begin{figure*}[t]
\centering
\includegraphics[width=.89\linewidth]{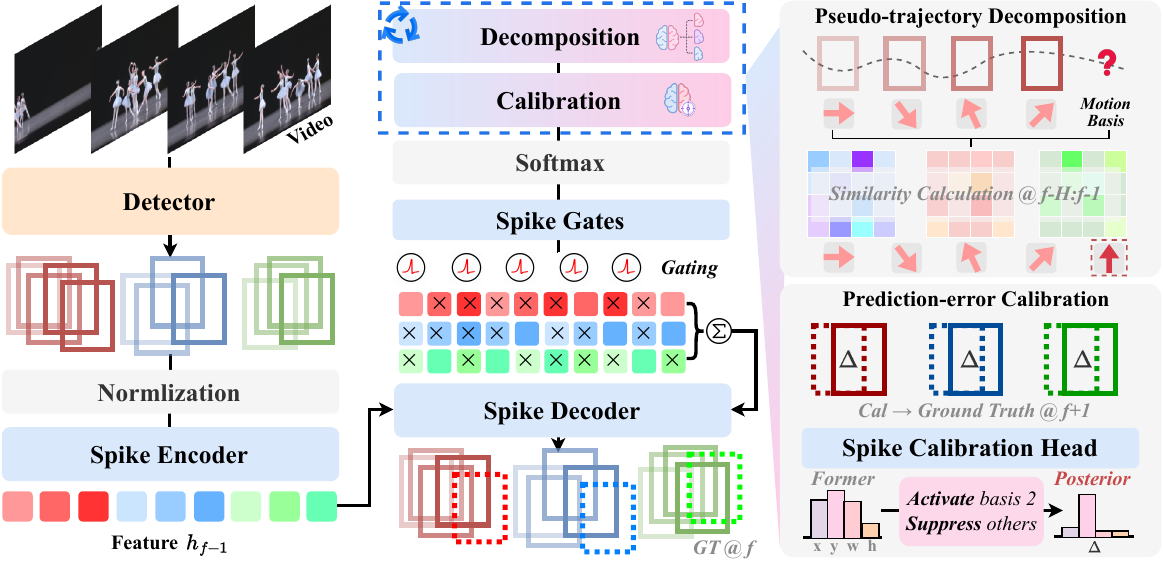} % Reduce the figure size so that it is slightly narrower than the column.
\caption{\textbf{Overview of SpikingMOT.} Given historical boxes of each active track, SpikingMOT encodes the recent trajectory, decomposes the motion state into pseudo-trajectory bases, selects informative bases through LIF spike firing, reconstructs a motion prior for association, and uses the matched detection residual to calibrate future basis responses.}
\label{fig:SpikingMOT_overview}
\end{figure*}

\subsection{SNNs in Vision Tasks}
In recent years, SNN methods have shown remarkable progress across a wide range of tasks, achieving performance comparable to that of ANNs~\cite{hu2023spiking}. Their applications span image classification~\cite{yao2023spiketransformer,yao2025scalingsdt}, object detection~\cite{luo2024spikeyolo,li2025yolo,liu2025sota,liu2021event,qu2025spiking}, semantic segmentation~\cite{lei2025spike2former}, video classification~\cite{zou2025spikevideoformer}, reinforcement learning~\cite{xu2025proxy,xu2026carebn}, temporal prediction~\cite{dong2025prednext} and object tracking~\cite{zhang2026spiketrack,shan2026sdtrack}. Existing SNN-based MOT methods either operate on event streams~\cite{wang2025spikemot} or retain a Kalman filter for motion prediction~\cite{zhong2026smtrack}. In contrast, our SpikingMOT introduces spike-driven motion prediction for RGB-based MOT and, together with SpikeYOLO, enables fully spike-driven tracking under the TBD paradigm.

\section{Method}\label{method}
SpikingMOT contains two task-specific components. Motion-basis decomposition constructs candidate displacements and performs spike-gated reconstruction, while prediction-error calibration updates their contributions for the next frame. Fig.~\ref{fig:SpikingMOT_overview} illustrates the overall framework.
\subsection{Problem Formulation}
Given a video sequence $\{V_f\}_{f=1}^{F}$ and an object detector $\mathcal{D}$, the detector produces candidate observations at frame $f$:
\begin{equation}
\mathcal{Z}_f
=
\mathcal{D}(V_f)
=
\{(\mathbf{z}_f^j,s_f^j)\}_{j=1}^{N_f},
\mathbf{z}_f^j
=
(c_{x,f}^j,c_{y,f}^j,w_f^j,h_f^j)^{\top},
\label{eq:detection_set}
\end{equation}
where $s_f^j$ is the detection confidence. We denote the state of track $\tau_i$ at frame $f$ by $\mathbf{x}_f^i=(c_{x,f}^i,c_{y,f}^i,w_f^i,h_f^i)^{\top}$. The goal is to output a trajectory set $\mathcal{T}_F=\{\tau_i\}_{i=1}^{N_T}$, where each tracklet stores its associated box states and identity label.

For each track $\tau_i\in\mathcal{T}_{f-1}$, SpikingMOT predicts the next box from its history and the calibrated basis posterior from the previous frame:
\begin{equation}
\hat{\mathbf{x}}_f^i
=
\Phi_{\Theta}
\left(
\mathbf{X}_{f-H:f-1}^{i};
\bar{\mathbf{a}}_{f-1}^{i}
\right),
\mathbf{X}_{f-H:f-1}^{i}
=
\{\mathbf{x}_{m}^{i}\}_{m=f-H}^{f-1},
\label{eq:online_prediction}
\end{equation}
where $\Phi_{\Theta}$ is the motion predictor and $H$ denotes the history length. For a track initialized at frame $f_i^{\rm init}$, we use the uniform
prior $\bar a_{f_i^{\rm init}-1,k}^{i}=1/K$ for $k=1,\ldots,K$.
where $K$ is the number of motion bases.

During training, the current prediction is supervised by the ground-truth box at frame $f$, while the calibrated posterior after frame $f$ is further required to improve the next-step prediction:
\begin{equation}
\begin{aligned}
\min_{\Theta,\mathbf{B}}\; \mathcal{L}_{\rm train}
=&
\frac{1}{|\mathcal{S}|}
\sum_{(i,f)\in\mathcal{S}}
\Big[
\ell\!\left(\hat{\mathbf{x}}_f^i,\mathbf{x}_f^i\right)
+
\alpha\,
\ell\!\left(\hat{\mathbf{x}}_{f+1}^{i,\mathrm{cal}},\mathbf{x}_{f+1}^{i}\right)
\Big]
\\
&+
\beta\mathcal{R}_{\rm spk}
+
\mu\mathcal{R}_{\rm div}.
\end{aligned}
\label{eq:training_objective}
\end{equation}
Here, $\mathcal{S}$ denotes trajectory segments with available states at frames $f$ and $f+1$. $\ell(\cdot)$ is the box prediction loss~\cite{rezatofighi2019generalized,girshick2014rich}. $\alpha$, $\beta$, and $\mu$ balance the loss terms. The spike sparsity and basis diversity regularizers are
\begin{equation}
\mathcal{R}_{\rm spk}
=
\frac{1}{2|\mathcal{S}|KT}
\sum_{(i,f)\in\mathcal{S}}
\sum_{q\in\{f,f+1\}}
\sum_{k=1}^{K}
\sum_{t=1}^{T}
 o_{q,k}^{i,t},
\label{eq:spike_regularization}
\end{equation}
\begin{equation}
\mathcal{R}_{\rm div}
=
\frac{1}{K(K-1)}
\sum_{\substack{k,m=1\\k\neq m}}^{K}
\left(
\frac{
\mathbf{b}_k^{\top}\mathbf{b}_m
}{
\|\mathbf{b}_k\|_2\|\mathbf{b}_m\|_2+\epsilon
}
\right)^2,
\label{eq:basis_diversity}
\end{equation}
where $\mathbf{B}=\{\mathbf{b}_k\}_{k=1}^{K}$ is the learnable motion-basis dictionary and $\epsilon>0$ is used for numerical stability. $\Theta$ collects the parameters of the trajectory encoder and all projection, displacement, and calibration heads.

\subsection{Trajectory-State Encoding}
Raw boxes contain scene-dependent coordinates and object scales. To obtain a transferable motion state, we encode relative box variations rather than absolute positions. For two consecutive boxes, the normalized displacement is
\begin{equation}
\mathbf{r}_{m}^{i}
=
\left(
\frac{\Delta c_{x,m}^{i}}{w_{m-1}^{i}},
\frac{\Delta c_{y,m}^{i}}{h_{m-1}^{i}},
\log\frac{w_m^i}{w_{m-1}^{i}},
\log\frac{h_m^i}{h_{m-1}^{i}}
\right)^{\top},
\label{eq:relative_motion}
\end{equation}
where $\Delta c_{x,m}^{i}=c_{x,m}^{i}-c_{x,m-1}^{i}$ and $\Delta c_{y,m}^{i}=c_{y,m}^{i}-c_{y,m-1}^{i}$. The recent motion sequence is encoded as
\begin{equation}
\mathbf{h}_f^i
=
E_{\Theta}
\left(
[\mathbf{r}_{f-H+1}^{i},\ldots,\mathbf{r}_{f-1}^{i}]
\right)
\in\mathbb{R}^{d_h},
\label{eq:motion_encoding}
\end{equation}
where $\mathbf{h}_f^i$ summarizes the motion state before frame $f$, and $E_{\Theta}$ denotes the trajectory-state encoder.

\subsection{Motion-Basis Decomposition}
\noindent\textbf{Basis-conditioned motion candidates.}
The ASP observation suggests that only a small subset of motion components is needed for each prediction. We first express this objective as sparse selection over a conceptual binary support $\mathbf{s}_f^i$:
\begin{equation}
\begin{aligned}
\min_{\mathbf{s}_f^i\in\{0,1\}^{K}}
&\;
\ell\left(
\mathbf{x}_{f-1}^{i}\oplus
\sum_{k=1}^{K}
\omega_{f,k}^{i}\boldsymbol{\delta}_{f,k}^{i},
\mathbf{x}_{f}^{i}
\right)
\\
\mathrm{s.t.}\quad
&1\leq\|\mathbf{s}_f^i\|_0\ll K,
\qquad
\omega_{f,k}^{i}
=
\frac{
 a_{f,k}^{i}s_{f,k}^{i}
}{
\sum_{m=1}^{K}a_{f,m}^{i}s_{f,m}^{i}
}.
\end{aligned}
\label{eq:sparse_basis_view}
\end{equation}
Here, $\boldsymbol{\delta}_{f,k}^{i}$ is the displacement proposed by basis $k$. In practice, the binary support is implemented by the effective spike gate introduced below.

Inspired by population coding~\cite{pouget2000information}, SpikingMOT decomposes the encoded motion state into $K$ basis-conditioned motion responses. The calibrated posterior from the previous frame is injected as a track-specific prior.
\begin{equation}
\boldsymbol{\delta}_{f,k}^{i}
=
P_{\Theta}
\left(
\mathbf{h}_f^i,\mathbf{b}_k
\right)
\in\mathbb{R}^{4}.
\label{eq:basis_displacement}
\end{equation}
Each basis produces a one-step candidate displacement, whose successive selection and propagation across frames form a pseudo-trajectory.

\noindent\textbf{Spike-gated reconstruction.}
The dense basis posterior is directly encoded as
$I_{f,k}^{i,t}=a_{f,k}^{i}$ for $t=1,\ldots,T$ and passed to a LIF neuron. The membrane dynamics are~\cite{neftci2019surrogate}
\begin{equation}
\begin{aligned}
\tilde{u}_{f,k}^{i,t}
&=
\lambda u_{f,k}^{i,t-1}+I_{f,k}^{i,t},
\quad
o_{f,k}^{i,t}=
\mathcal{H}\!\left(\tilde{u}_{f,k}^{i,t}-\vartheta\right),
\\
u_{f,k}^{i,t}
&=
\tilde{u}_{f,k}^{i,t}-\vartheta o_{f,k}^{i,t}.
\end{aligned}
\label{eq:lif_dynamics}
\end{equation}
Here, $u$ and $\tilde u$ are the post-reset and pre-reset membrane potentials, $\lambda$ is the leakage factor, $\vartheta$ is the firing threshold, and $o_{f,k}^{i,t}\in\{0,1\}$ is the emitted spike. During backpropagation, the derivative of $\mathcal{H}$ is approximated using a surrogate gradient \textit{\underline{(cf. Supp D.1)}}.

The emitted spikes are aggregated into a bounded gate:
\begin{equation}
g_{f,k}^{i}
=
\frac{1}{D}
\operatorname{clip}
\left(
\sum_{t=1}^{T}o_{f,k}^{i,t},
0,D
\right),
\qquad
g_{f,k}^{i}\in[0,1],
\label{eq:spike_gate}
\end{equation}
where $D$ is maximum integer value emitted during training.
The effective basis weights and reconstructed motion are
\begin{equation}
\begin{aligned}
\omega_{f,k}^{i}
&=
\frac{
a_{f,k}^{i}\tilde g_{f,k}^{i}
}{
\sum_{m=1}^{K}a_{f,m}^{i}\tilde g_{f,m}^{i}
},\\
\widehat{\boldsymbol{\Delta}}_f^i
&=
\sum_{k=1}^{K}
\omega_{f,k}^{i}\boldsymbol{\delta}_{f,k}^{i},
\qquad
\hat{\mathbf{x}}_f^i
=
\mathbf{x}_{f-1}^{i}
\oplus
\widehat{\boldsymbol{\Delta}}_f^i.
\end{aligned}
\label{eq:motion_reconstruction}
\end{equation}
The predicted boxes are matched with detections using Hungarian assignment~\cite{kuhn1955hungarian}. We follow the association cost, confidence filtering, and matching thresholds of MambaTrack.

\subsection{Prediction-Error Calibration}
Following the predictive-coding view that residual errors provide corrective signals for internal predictions~\cite{rao1999predictive}, SpikingMOT uses the matched detection residual to recalibrate the basis posterior for future motion prediction. For a matched pair $(\tau_i,\mathbf{z}_f^j)$, we compute a scale-normalized box residual:
\begin{equation}
\mathbf{e}_f^i
=
\left(
\frac{c_{x,f}^{j}-\hat{c}_{x,f}^{i}}{\hat{w}_f^i},
\frac{c_{y,f}^{j}-\hat{c}_{y,f}^{i}}{\hat{h}_f^i},
\log\frac{w_f^j}{\hat{w}_f^i},
\log\frac{h_f^j}{\hat{h}_f^i}
\right)^{\top}.
\label{eq:prediction_error}
\end{equation}
During training, the observation in Eq.~\eqref{eq:prediction_error} is the ground-truth box with known identity correspondence. During inference, it is the detection obtained through Hungarian assignment. The calibration head predicts a logit correction for all bases:
\begin{equation}
\Delta\boldsymbol{\eta}_f^i
=
C_{\Theta}
\left(
[\mathbf{h}_f^i;\mathbf{e}_f^i;\mathbf{a}_f^i;\tilde{\mathbf{g}}_f^i]
\right)
\in\mathbb{R}^{K}.
\label{eq:calibration_head}
\end{equation}
The calibrated posterior is
\begin{equation}
\bar{\mathbf{a}}_f^i
=
\operatorname{softmax}
\left(
\log(\mathbf{a}_f^i+\epsilon)
+
\Delta\boldsymbol{\eta}_f^i
\right).
\label{eq:error_calibration}
\end{equation}
For an unmatched active track, no error correction is applied and we set $\bar{\mathbf a}_f^i=\mathbf a_f^i$. The calibrated posterior is carried to the next prediction step:
\begin{equation}
\hat{\mathbf{x}}_{f+1}^{i,\mathrm{cal}}
=
\Phi_{\Theta}
\left(
\mathbf{X}_{f-H+1:f}^{i};
\bar{\mathbf{a}}_{f}^{i}
\right).
\label{eq:calibrated_next_prediction}
\end{equation}
Thus, the residual observed at frame $f$ updates the basis prior used to predict frame $f+1$, forming a closed loop between motion prediction and error-driven basis selection \textit{\underline{(cf. Supp B)}}.

\begin{table*}[t]
\centering
\resizebox{\linewidth}{!}{%
\setlength{\tabcolsep}{4.2pt}
\begin{tabular}{c|l|ccc|cccc|cccc}
\toprule
\multirow{2}{*}{} 
& \multirow{2}{*}{Tracker}
& \multicolumn{3}{c|}{Efficiency}
& \multicolumn{4}{c|}{DanceTrack}
& \multicolumn{4}{c}{SportsMOT}
\\
\cmidrule{3-13}
&
& Para. & Eng. & T$\times$D
& HOTA$\uparrow$ & IDF1$\uparrow$ & AssA$\uparrow$
& MOTA$\uparrow$ & HOTA$\uparrow$ & IDF1$\uparrow$
& AssA$\uparrow$ & MOTA$\uparrow$
\\
\midrule

\multirow{5}{*}{\rotatebox{90}{KF}}
& *SMTrack-SY$_{23M}$
& N/A & N/A & N/A
& 37.7 & 38.5 & 27.5 & 70.6
& 43.3 & 48.9 & 37.6 & 66.5 \\

& SMTrack-SY$_{69M}$
& N/A & N/A & N/A
& 42.7 & 43.4 & 27.8 & 74.6
& 57.5 & 56.2 & 43.9 & 82.8 \\

& FairMOT
& N/A & N/A & N/A
& 39.7 & 40.8 & 23.8 & 82.2
& 49.3 & 53.5 & 34.7 & 86.4 \\

& ByteTrack
& N/A & N/A & N/A
& 47.7 & 53.9 & 32.1 & 89.6
& 63.4 & 70.3 & 51.3 & 95.7 \\

& OC-SORT
& N/A & N/A & N/A
& 55.1 & 54.6 & 38.3 & \underline{92.0}
& 73.7 & 74.0 & 61.5 & 96.5 \\

\midrule

\multirow{9}{*}{\rotatebox{90}{ANN}}
& QDTrack
& N/A & N/A & N/A
& 54.2 & 50.4 & 36.8 & 87.7
& 60.4 & 62.3 & 47.2 & 90.1 \\

& GTR
& N/A & N/A & N/A
& 48.0 & 50.3 & 31.9 & 84.7
& 54.5 & 55.8 & 45.9 & 67.9 \\

& TransTrack
& N/A & N/A & N/A
& 45.5 & 45.2 & 27.5 & 88.4
& 68.9 & 71.5 & 57.5 & 92.6 \\

& MotionTrack
& 25.2 & 55.8 & N/A
& - & - & - & -
& 74.0 & 74.0 & 61.7 & \underline{96.6} \\

& *SportMamba
& 19.3 & 32.2 & N/A
& 51.8 & 49.5 & 37.5 & 88.7
& 74.3 & 74.7 & \textbf{63.8} & \textbf{96.9} \\

& DiffMOT
& 14.6 & 21.9 & N/A
& \underline{56.1} & \underline{55.7} & 38.5 & \textbf{92.2}
& 72.1 & 72.8 & 60.5 & 94.5 \\

& MambaMOT+
& 6.5 & 19.1 & N/A
& \underline{56.1} & 54.9 & 39.0 & 90.3
& 71.3 & 71.1 & 58.6 & 94.9 \\

& MambaTrack
& 13.3 & 26.9 & N/A
& 55.8 & \textbf{56.8} & \underline{39.8} & 90.1
& 72.6 & 72.8 & 60.3 & 95.3 \\

& *TrackSSM
& 5.1 & 17.4 & N/A
& 52.7 & 53.5 & \textbf{41.0} & 88.2
& \underline{74.4} & 74.5 & 62.4 & 95.3 \\

\midrule

\multirow{7}{*}{\rotatebox{90}{SNN (ours)}}
& \multirow{2}{*}{SpikingMOT-SY$_{23M}$}
& 3.6 & 2.2 & 4$\times$1
& \cellcolor{gray!10}40.2 & \cellcolor{gray!10}40.1 & \cellcolor{gray!10}26.2 & \cellcolor{gray!10}73.1
& \cellcolor{gray!10}48.8 & \cellcolor{gray!10}50.4 & \cellcolor{gray!10}39.7 & \cellcolor{gray!10}72.3 \\
& 
& 3.6 & 1.6 & 1$\times$1
& \cellcolor{gray!10}41.5 & \cellcolor{gray!10}42.6 & \cellcolor{gray!10}27.6 & \cellcolor{gray!10}73.8
& \cellcolor{gray!10}49.4 & \cellcolor{gray!10}51.3 & \cellcolor{gray!10}40.6 & \cellcolor{gray!10}72.8 \\

\cmidrule{2-13}

& \multirow{2}{*}{SpikingMOT-SY$_{69M}$}
& 3.6 & 3.2 & 4$\times$1
& \cellcolor{gray!10}45.2 & \cellcolor{gray!10}45.0 & \cellcolor{gray!10}30.4 & \cellcolor{gray!10}76.6
& \cellcolor{gray!10}54.7 & \cellcolor{gray!10}54.7 & \cellcolor{gray!10}45.1 & \cellcolor{gray!10}89.1 \\
&
& 3.6 & 2.3 & 1$\times$1
& \cellcolor{gray!10}46.6 & \cellcolor{gray!10}47.1 & \cellcolor{gray!10}32.2 & \cellcolor{gray!10}79.1
& \cellcolor{gray!10}55.4 & \cellcolor{gray!10}56.7 & \cellcolor{gray!10}46.4 & \cellcolor{gray!10}89.1 \\

\cmidrule{2-13}

& \multirow{2}{*}{SpikingMOT-YL$_{104M}$}
& \textbf{3.6} & \underline{5.4} & 4$\times$1
& \cellcolor{gray!10}53.7 & \cellcolor{gray!10}52.7 & \cellcolor{gray!10}36.1 & \cellcolor{gray!10}91.1
& \cellcolor{gray!10}\textbf{74.9} & \cellcolor{gray!10}\underline{75.8} & \cellcolor{gray!10}63.6 & \cellcolor{gray!10}95.4 \\
&
& \textbf{3.6} & \textbf{3.6} & 1$\times$1
& \cellcolor{gray!10}\textbf{56.5} & \cellcolor{gray!10}\underline{55.7} & \cellcolor{gray!10}\underline{39.8} & \cellcolor{gray!10}91.2
& \cellcolor{gray!10}\textbf{74.9} & \cellcolor{gray!10}\textbf{75.9} & \cellcolor{gray!10}\underline{63.7} & \cellcolor{gray!10}95.4 \\

\bottomrule
\end{tabular}}
\caption{\textbf{Comparison of performance and efficiency across tracking methods.}
\textit{Para.} and \textit{Eng.} denote parameters (M) and energy (mJ),
respectively; $T$ and $D$ represent the timestep and maximum integer value emitted during training.
Results marked with $^{*}$ are reproduced by us. SY means SpikeYOLO, YL means YOLOX. All metrics are reported in percentage (\%). \textit{Para.} counts only the learnable parameters of the motion predictor, while \textit{Eng.} reports per-frame energy under the corresponding detector. The top two comparable results are highlighted in \textbf{bold} and \underline{underlined}, respectively. Same for subsequent tables.}
\label{tab1}
\end{table*}

\begin{table*}[t]
\centering
\resizebox{\linewidth}{!}{%
\setlength{\tabcolsep}{4.2pt}
\begin{tabular}{c|l|ccc|cccc|cccc}
\toprule
\multirow{2}{*}{}
& \multirow{2}{*}{Tracker}
& \multicolumn{3}{c|}{Efficiency}
& \multicolumn{4}{c|}{SeaDroneSee}
& \multicolumn{4}{c}{MOT17}
\\
\cmidrule{3-13}
&
& Para. & Eng. & T$\times$D
& HOTA$\uparrow$ & IDF1$\uparrow$ & MOTP$\uparrow$ & MOTA$\uparrow$
& HOTA$\uparrow$ & IDF1$\uparrow$ & AssA$\uparrow$ & MOTA$\uparrow$
\\

\midrule

\multirow{6}{*}{\rotatebox{90}{KF}}
& DeepSORT w/ ECC
& N/A & N/A & N/A
& 66.6 & 77.2 & 18.8 & 80.0
& - & - & - & - \\

& ByteTrack
& N/A & N/A & N/A
& 65.0 & 77.3 & 20.9 & 76.9
& 62.8 & 76.8 & 62.1 & 78.7 \\

& DeepSORT
& N/A & N/A & N/A
& 62.7 & 72.0 & 20.1 & 77.2
& \textbf{63.8} & \textbf{78.3} & \underline{62.2} & \textbf{80.3} \\

& OC-SORT
& N/A & N/A & N/A
& 60.8 & 69.2 & 18.5 & 72.4
& \underline{63.2} & \underline{77.5} & \textbf{63.4} & 78.0 \\

& FairMOT R34
& N/A & N/A & N/A
& - & 40.8 & 27.3 & 30.5
& - & - & - & - \\

& FairMOT D34
& N/A & N/A & N/A
& - & 43.8 & 20.9 & 36.5
& 59.3 & 72.3 & 58.0 & 73.7 \\

\midrule

\multirow{3}{*}{\rotatebox{90}{ANN}}
& *MambaTrack
& 13.3 & 26.9 & N/A
& 68.5 & 76.8 & 36.2 & 79.8
& 60.1 & 72.9 & 58.4 & 77.1 \\

& Tracktor
& 60.9 & 111.3 & N/A
& 46.1 & 49.9 & 21.4 & 47.8
& 52.2 & 64.7 & 51.0 & 67.8 \\

& Tracktor++
& 87.0 & 114.4 & N/A
& - & \textbf{80.5} & 20.1 & 71.9
& - & - & - & - \\

\midrule

\multirow{7}{*}{\rotatebox{90}{SNN (ours)}}
& \multirow{2}{*}{SpikingMOT-SY$_{23M}$}
& 3.6 & 2.2 & 4$\times$1
& \cellcolor{gray!10}50.3 & \cellcolor{gray!10}55.7 & \cellcolor{gray!10}23.8 & \cellcolor{gray!10}52.5
& \cellcolor{gray!10}52.9 & \cellcolor{gray!10}64.0 & \cellcolor{gray!10}51.0 & \cellcolor{gray!10}69.2 \\
&
& 3.6 & 1.6 & 1$\times$1
& \cellcolor{gray!10}50.5 & \cellcolor{gray!10}55.9 & \cellcolor{gray!10}23.9 & \cellcolor{gray!10}52.8
& \cellcolor{gray!10}52.7 & \cellcolor{gray!10}63.9 & \cellcolor{gray!10}50.8 & \cellcolor{gray!10}69.1 \\

\cmidrule{2-13}

& \multirow{2}{*}{SpikingMOT-SY$_{69M}$}
& 3.6 & 3.2 & 4$\times$1
& \cellcolor{gray!10}59.7 & \cellcolor{gray!10}64.2 & \cellcolor{gray!10}31.9 & \cellcolor{gray!10}61.8
& \cellcolor{gray!10}54.1 & \cellcolor{gray!10}61.2 & \cellcolor{gray!10}49.9 & \cellcolor{gray!10}72.2 \\
&
& 3.6 & 2.3 & 1$\times$1
& \cellcolor{gray!10}59.8 & \cellcolor{gray!10}64.3 & \cellcolor{gray!10}32.4 & \cellcolor{gray!10}62.0
& \cellcolor{gray!10}53.8 & \cellcolor{gray!10}61.0 & \cellcolor{gray!10}49.9 & \cellcolor{gray!10}72.1 \\

\cmidrule{2-13}

& \multirow{2}{*}{SpikingMOT-YL$_{104M}$}
& \textbf{3.6} & \underline{5.4} & 4$\times$1
& \cellcolor{gray!10}\textbf{69.1} & \cellcolor{gray!10}\underline{79.4} & \cellcolor{gray!10}\textbf{38.8} & \cellcolor{gray!10}\underline{82.1}
& \cellcolor{gray!10}61.4 & \cellcolor{gray!10}74.4 & \cellcolor{gray!10}59.5 & \cellcolor{gray!10}\underline{78.9} \\
&
& \textbf{3.6} & \textbf{3.6} & 1$\times$1
& \cellcolor{gray!10}\underline{69.0} & \cellcolor{gray!10}78.8 & \cellcolor{gray!10}\underline{38.2} & \cellcolor{gray!10}\textbf{82.6}
& \cellcolor{gray!10}61.3 & \cellcolor{gray!10}74.3 & \cellcolor{gray!10}59.0 & \cellcolor{gray!10}\underline{78.9} \\

\bottomrule
\end{tabular}}
\caption{\textbf{Comparison of performance and efficiency across tracking methods.}
\textit{Para.} and \textit{Eng.} denote parameters (M) and energy (mJ),
respectively. SY means SpikeYOLO, YL means YOLOX. All metrics are reported in percentage (\%).}
\label{tab2}
\end{table*}

\begin{figure*}[t]
\centering
\includegraphics[width=\linewidth]{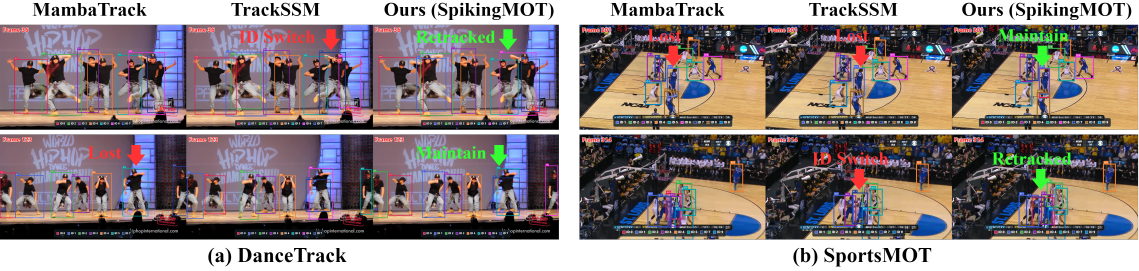} % Reduce the figure size so that it is slightly narrower than the column.
\caption{Qualitative comparison of SpikingMOT in DanceTrack and SportsMOT dataset.}
\label{fig3}
\end{figure*}

\section{Experiments}\label{exp}
\subsection{Implementation Details}
\textbf{Metrics.} Following standard evaluation protocols, we report HOTA~\cite{luiten2021hota}, MOTA~\cite{bernardin2008clearmot}, IDF1~\cite{ristani2016performance}, AssA, MOTP, and Energy \textit{\underline{(cf. Supp C)}}. HOTA serves as the metric by jointly assessing detection and association quality. MOTA summarizes false positives, false negatives, and identity switches, while IDF1 evaluates identity preservation over complete trajectories. AssA isolates the accuracy of associations, and MOTP measures bounding-box localization precision.

\textbf{Training.} All spiking layers employ soft-reset LIF neurons, whose membrane states are initialized independently for each tracklet \textit{\underline{(cf. Supp D.1, D.2)}}. We use a history length of $H=5$ and $K=8$ pseudo-trajectory bases. The loss weights are set to $\alpha=1.0$,
$\beta=10^{-3}$, $\mu=10^{-2}$, and $\epsilon=10^{-6}$ \textit{\underline{(cf. Supp D.3)}}.  For fair comparison, the training schedules and association hyperparameters follow MambaTrack~\cite{xiao2024mambatrack} and TrackSSM~\cite{hu2026trackssm}. Since the publicly available SpikeYOLO~\cite{luo2024spikeyolo} models were not trained on SportsMOT or SeaDroneSee, we initialize them from COCO-pretrained~\cite{lin2014coco} weights and fine-tune the SY$_{23M}$ and SY$_{69M}$ variants separately on each dataset. The detection head is adapted to the target category, while the original SpikeYOLO optimization and augmentation protocol is retained. The detector and motion predictor are trained separately, and the best validation checkpoint is used for tracking evaluation. All experiments are conducted on a single NVIDIA RTX 3080Ti GPU.

\subsection{Benchmark Results}
Tables~\ref{tab1} and~\ref{tab2} compare SpikingMOT with representative KF-based~\cite{zhong2026smtrack,zhang2022bytetrack}, ANN-based~\cite{pang2021qdtrack,zhou2022gtr,sun2020transtrack,xiao2024motiontrack,khanna2025sportmamba,huang2025mambamot,hu2026trackssm}, and spike-driven trackers across four benchmarks with substantially different motion characteristics.

\textbf{DanceTrack}~\cite{sun2022dancetrack} is association-sensitive because dancers exhibit similar appearances, frequent crossings, and highly non-linear movements. SpikingMOT achieves the strongest HOTA, although it does not obtain the highest MOTA. This discrepancy mainly arises from more reliable temporal association rather than simply retaining more detections. By representing motion through selectively activated pseudo-trajectory bases, SpikingMOT can adapt between different local motion patterns instead of forcing rapidly changing trajectories into a dense transition model.

\textbf{SportsMOT}~\cite{cui2023sportsmot} contains rapid acceleration, abrupt direction changes, and visually similar players, making motion prediction a central bottleneck. SpikingMOT achieves the best overall balance of HOTA, IDF1, and AssA, with its advantage becoming more evident when paired with a stronger detector. As localization becomes reliable, association increasingly depends on the ability to represent motion hypotheses, which is precisely addressed by SpikingMOT through brain-inspired strategies.

\textbf{SeaDroneSee}~\cite{varga2022seadronessee} presents distinct challenges, including aerial viewpoints, small targets, and camera-induced motion. SpikingMOT remains strong in both tracking and localization, indicating that prediction error provides an effective online reliability signal for selecting appropriate motion dynamics under uncertain observations.

\textbf{MOT17}~\cite{dendorfer2021motchallenge} is dominated by crowded pedestrian scenes and appearance ambiguity, while pedestrian motion is generally more regular. The result clarifies that adaptive sparse motion modeling is most effective when complex dynamics dominate association errors, but offers less benefit under prolonged occlusion and severe appearance ambiguity. Nevertheless, SpikingMOT achieves comparable performance with substantially lower model complexity and energy consumption, demonstrating its general applicability beyond highly dynamic scenarios.

\begin{table*}[t]
\centering
\small
\setlength{\tabcolsep}{3pt}
\renewcommand{\arraystretch}{1.08}

% -------------------------------------------------
% (a) Sparse ANN alternatives
% -------------------------------------------------
\begin{subtable}[t]{0.27\textwidth}
\centering
\caption{\textbf{Sparse ANN alternatives.}}
\label{tab:ablation_sparse_ann}
\resizebox{\linewidth}{!}{%
\begin{tabular}{l|cc}
\toprule
Tracker
& HOTA$\uparrow$
& MOTA$\uparrow$ \\
\midrule
Dense ANN
& 72.6
& 95.3 \\
+ Dropout
& 73.6
& 95.2 \\
+ $\ell_1$ Sparsity
& 72.9
& 95.1 \\
+ Top-$k$ Pruning
& 73.1
& 95.2 \\
\rowcolor{gray!10}
\textbf{SpikingMOT}
& \textbf{74.9}
& \textbf{95.4} \\
\bottomrule
\end{tabular}%
}
\end{subtable}
\hfill
% -------------------------------------------------
% (b) Core components
% -------------------------------------------------
\begin{subtable}[t]{0.3\textwidth}
\centering
\caption{\textbf{Core components.}}
\label{tab:ablation_components}
\resizebox{\linewidth}{!}{%
\begin{tabular}{ccc|cc}
\toprule
\multicolumn{3}{c|}{Module}
& \multirow{2}{*}{HOTA$\uparrow$}
& \multirow{2}{*}{MOTA$\uparrow$} \\
\cmidrule(lr){1-3}
Spike
& Decomp.
& Calib.
&  &  \\
\midrule
&
&
&
72.6
& 95.3 \\
\checkmark
&
&
&
73.3
& 95.2 \\
\checkmark
& \checkmark
&
&
74.1
& 95.3 \\
\rowcolor{gray!10}
\checkmark
& \checkmark
& \checkmark
& \textbf{74.9}
& \textbf{95.4} \\
\bottomrule
\end{tabular}%
}
\end{subtable}
\hfill
% -------------------------------------------------
% (c) Basis number
% -------------------------------------------------
\begin{subtable}[t]{0.183\textwidth}
\centering
\caption{\textbf{Basis number.}}
\label{tab:ablation_basis}
\resizebox{\linewidth}{!}{%
\begin{tabular}{c|ccc}
\toprule
$K$
& Eng.
& SR
& HOTA$\uparrow$ \\
\midrule
1
& 3.1
& 22.6
& 73.3 \\
4
& 3.5
& 17.8
& 74.6 \\
\rowcolor{gray!10}
8
& \textbf{3.6}
& \textbf{15.4}
& \textbf{74.9} \\
16
& 3.9
& 16.1
& 74.7 \\
32
& 4.4
& 18.7
& 74.3 \\
\bottomrule
\end{tabular}%
}
\end{subtable}
\hfill
% -------------------------------------------------
% (d) Spiking neuron types
% -------------------------------------------------
\begin{subtable}[t]{0.195\textwidth}
\centering
\caption{\textbf{Neuron types.}}
\label{tab:neuron_types}

{\renewcommand{\arraystretch}{1.35}%
\resizebox{\linewidth}{!}{%
\begin{tabular}{l|cc}
\toprule
Neuron
& HOTA$\uparrow$
& MOTA$\uparrow$ \\
\midrule
IF
& 73.6
& 95.1 \\
\rowcolor{gray!10}
LIF
& \textbf{74.9}
& \textbf{95.4} \\
PLIF
& 74.6
& 95.3 \\
ALIF
& 74.2
& 95.2 \\
\bottomrule
\end{tabular}%
}}

\end{subtable}

\caption{\textbf{Ablation studies on SportsMOT.}}
\label{tab:ablation_four_panels}
\end{table*}

\begin{table}[t]
\centering
\noindent
\begin{minipage}[t]{0.61\linewidth}
\vspace{0pt}
\centering
\vspace{-1mm}
\setlength{\tabcolsep}{2.2pt}
\renewcommand{\arraystretch}{1.12}
\resizebox{\linewidth}{!}{
\begin{tabular}{l|l|cc}
\toprule
Detector
& Tracker & MOTA$\uparrow$ & IDF1$\uparrow$ \\
\midrule
\multirow{3}{*}{SY$_{23M}$}
& KF & 63.5 & 44.7 \\
& MambaTrack & 66.1 & 50.5 \\
& \cellcolor{gray!10}\textbf{SpikingMOT} & \cellcolor{gray!10}\textbf{72.8} & \cellcolor{gray!10}\textbf{51.3} \\
\midrule
\multirow{3}{*}{SY$_{69M}$}
& KF & 83.2 & 45.5 \\
& MambaTrack & 88.4 & 55.9 \\
& \cellcolor{gray!10}\textbf{SpikingMOT} & \cellcolor{gray!10}\textbf{89.1} & \cellcolor{gray!10}\textbf{56.7} \\
\midrule
\multirow{3}{*}{YL$_{104M}$}
& KF & 92.0 & 71.0 \\
& MambaTrack & 95.3 & 72.8 \\
& \cellcolor{gray!10}\textbf{SpikingMOT} & \cellcolor{gray!10}\textbf{95.4} & \cellcolor{gray!10}\textbf{75.9} \\
\bottomrule
\end{tabular}}
\caption{Detector-wise tracking performance of SpikingMOT over KF and MambaTrack on SportsMOT.}
\label{tab:ablation_detector_gain}
\end{minipage}
\hfill
\begin{minipage}[t]{0.35\linewidth}
% \vspace{0pt}
\vspace{-1mm}
\centering
\includegraphics[width=1\linewidth,keepaspectratio]{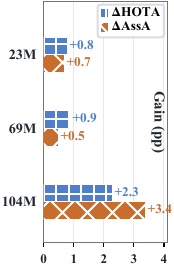}
\vspace{-5mm}
\captionof{figure}{Detector-wise gain over MambaTrack.}
\label{fig:detector_gain_strip}
\end{minipage}
% \vspace{-3.5mm}
\end{table}

\subsection{Ablation Study}
\textbf{Sparse ANN alternatives.}
Table~\ref{tab:ablation_sparse_ann} shows that sparsity alone cannot explain the advantage of SpikingMOT. Conventional strategies suppress responses through random masking or parameter shrinkage, without considering the current motion state. In contrast, SpikingMOT adapts its firing pattern to the evolving trajectory, indicating that performance depends not on how many responses are removed, but on whether the retained responses match the current dynamics.

\textbf{Core components.} 
Table~\ref{tab:ablation_components} clarifies the complementary roles of the proposed components. Pseudo-trajectory decomposition provides diverse motion hypotheses, spiking dynamics select relevant ones, and prediction-error calibration updates this selection using online observations. Together, they form a closed-loop sparse motion inference process.

\textbf{Number of motion bases.}
Table~\ref{tab:ablation_basis} reveals a trade-off between motion diversity and basis redundancy. \textit{SR} denotes the mean spike firing rate. Too few bases limit expressiveness, whereas too many introduce overlapping hypotheses and unnecessary firing. A moderate basis set therefore provides sufficient motion diversity while preserving selective and energy-efficient activation.

\textbf{Spiking neuron types.}
Table~\ref{tab:neuron_types} shows that neuron dynamics should remain sufficiently simple for motion prediction. IF lacks temporal leakage, while PLIF~\cite{fang2021incorporating} and ALIF~\cite{bellec2018long} introduce additional adaptive parameters that may overfit limited trajectory histories. LIF provides a better balance by retaining short-term motion memory through fixed leakage without adding unnecessary optimization complexity.

\textbf{Generalization across detectors.} Table~\ref{tab:ablation_detector_gain} and Fig.~\ref{fig:detector_gain_strip} evaluate different motion predictors on SportsMOT while fixing the detector outputs and association protocol. SpikingMOT consistently outperforms the Kalman filter and MambaTrack across detectors, showing that its gains are not detector-specific. This suggests that as detection noise decreases, motion modeling becomes a more critical bottleneck. By selectively activating informative motion bases and adapting them through prediction errors, SpikingMOT better converts accurate detections into reliable identity associations.

\begin{figure}[t]
\centering
\includegraphics[width=\linewidth,height=54mm]{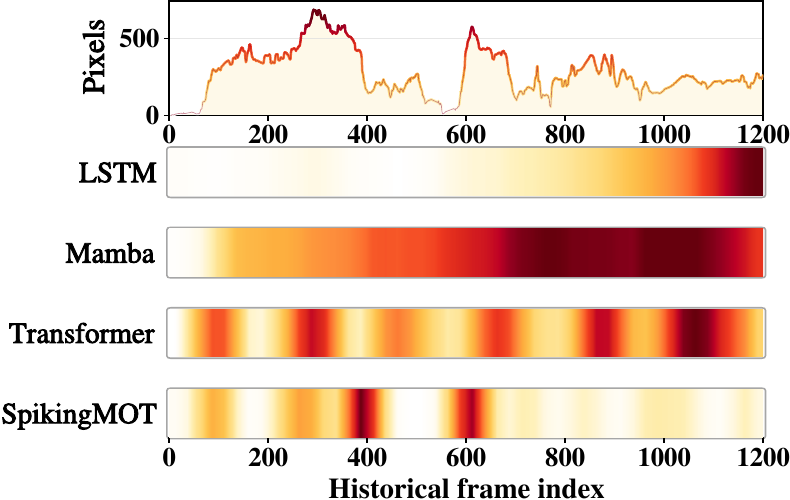} % Reduce the figure size so that it is slightly narrower than the column.
\caption{\textbf{Temporal influence of different motion models.}
The upper curve shows the historical displacement in DanceTrack0005, and the lower heatmaps indicate the contribution of each historical frame to the current prediction.}
\label{fig4}
\end{figure}

\textbf{Temporal influence analysis.} Given a track history, the reference prediction is obtained from the original sequence. For each historical frame $\tau$, we replace only its observation with temporal interpolation from the remaining history, rerun the model, and measure the influence of $\tau$ by the $\ell_2$ distance between the reference and perturbed predictions. The resulting influence values are normalized within each model. As shown in Fig.~\ref{fig4}, LSTM mainly depends on recent frames, Mamba distributes influence broadly, and Transformer exhibits scattered responses. In contrast, SpikingMOT forms a sparse yet non-myopic pattern, focusing on salient motion changes while suppressing redundant observations.

\textbf{Qualitative results.} As shown in Fig.~\ref{fig3}, SpikingMOT maintains more consistent identities under crowded interactions, abrupt motion changes, and partial occlusions on both benchmarks. Compared with competing trackers, it produces fewer identity switches and more stable trajectories when appearance cues become ambiguous \textit{\underline{(cf. Supp D.4)}}.

\section{Conclusion}
This paper investigates the ASP phenomenon in MOT trajectory prediction. We formulate sparse motion modeling through SNNs and propose SpikingMOT, a spike-driven tracker that refines trajectory prediction via motion-basis decomposition and prediction-error calibration. Across diverse benchmarks, SpikingMOT demonstrates significant performance improvement over ANN baselines with superior efficiency. Our results suggest that MOT trackers should treat sparsity as a learnable principle for motion reasoning. One limitation is that energy consumption is estimated from operation counts rather than measured on neuromorphic hardware. Future work will pursue deployment on neuromorphic chips~\cite{pei2019towards} and evaluation with spike-camera data~\cite{huang2023faster} in real-world scenarios.

% \newpage
\section{Acknowledgments}

\bibliography{aaai2027}


\end{document}